\documentclass[runningheads]{llncs}
\usepackage{graphicx}
\usepackage{algorithm}
\usepackage[noend]{algpseudocode}
\usepackage{cite}

\usepackage{amsfonts}
\usepackage{comment}
\usepackage{amsmath}
\usepackage{color}

\usepackage{enumitem}
\usepackage{hyperref}
\usepackage[normalem]{ulem}

\usepackage{xcolor}
\usepackage{soul}
\sethlcolor{black}
\sethlcolor{black}
\usepackage{multirow}
\usepackage{hyperref}

\begin{document}
\title{Hyperparameter optimization in deep multi-target prediction}
%
%
\author{Dimitrios Iliadis\inst{1}\orcidID{0000-0002-3676-5940} \and
Marcel Wever\inst{2}\orcidID{0000-0001-9782-6818} \and
Bernard De Baets\inst{1}\orcidID{0000-0002-3876-620X} \and
Willem Waegeman\inst{1}\orcidID{0000--0002-5950-3003}}

\institute{KERMIT, Department of Data Analysis and Mathematical Modelling, Ghent University, Coupure links 653, B-9000 Ghent, Belgium\\
\email{\{dimitrios.iliadis,bernard.debaets,willem.waegeman\}@ugent.be} \and
Department of Computer Science, Ludwig-Maximilians-University Munich, Akademiestr. 7, 80799 Munich,  Germany \\
\email{marcel.wever@ifi.lmu.de}\\
}

\maketitle              
\begin{abstract}
As a result of the ever increasing complexity of configuring and fine-tuning machine learning models, the field of automated machine learning (AutoML) has emerged over the past decade. However, software implementations like Auto-WEKA and Auto-sklearn typically focus on classical machine learning (ML) tasks such as classification and regression. Our work can be seen as the first attempt at offering a single AutoML framework for most problem settings that fall under the umbrella of multi-target prediction, which includes popular ML settings such as multi-label classification, multivariate regression, multi-task learning, dyadic prediction, matrix completion, and zero-shot learning. Automated problem selection and model configuration are achieved by extending DeepMTP, a general deep learning framework for MTP problem settings, with popular hyperparameter optimization (HPO) methods. Our extensive benchmarking across different datasets and MTP problem settings identifies cases where specific HPO methods outperform others.

\end{abstract}
\section{Introduction}
The past decade of AI research has been dominated by significant advances in deep learning. From convolutional neural networks (CNNs)~\cite{krizhevsky2012imagenet,simonyan2014very} to generative adversarial networks (GANs)~\cite{goodfellow2014generative} and transformers~\cite{vaswani2017attention}, deep learning architectures have enabled major breakthroughs in several application areas, such as computer vision, speech recognition, and protein folding~\cite{jumper2021highly}. However, the configuration of these increasingly complex architectures requires multiple design decisions that are not standardized. Selecting the appropriate architecture and hyperparameters for the optimal neural network is usually reserved for highly experienced users who navigate the configuration space through trial and error. This process becomes dull in a per-dataset case and essentially infeasible when one wants to offer a model in a software tool that thousands of inexperienced users will potentially use.

The increased demand for machine learning applications and the limited availability of expertise has led to the emergence of the field of automated machine learning (AutoML) \cite{DBLP:books/sp/HKV2019}, which is concerned with automating the process of engineering machine learning applications.
In particular, this field aims to develop methods that help to move away from the tedious task of manually configuring machine learning algorithms to a data-driven approach that is able to efficiently navigate through  the space of potential solution candidates while maintaining near-optimal performance.
An essential sub-task that needs to be tackled to achieve this performance deals with the optimization of hyperparameters. Since we focus on hyperparameter optimization (HPO) in this work, we refer to several AutoML-related surveys~\cite{he2021automl,elsken2019neural,DBLP:journals/corr/abs-2107-05847} for a more thorough introduction and overview.
Especially in the case of neural networks, hyperparameter optimization plays a significant role as the hyperparameters greatly affect the computational complexity and the generalization performance.

A subarea of AutoML research, also known as neural architecture search (NAS)~\cite{elsken2019neural}, is solely concerned with hyperparameter optimization of a particular class of models, namely neural networks. While NAS has demonstrated promising performance~\cite{zoph2016neural} and efficiency~\cite{pham2018efficient} improvements, most existing work has focused on the challenging, yet narrow task of image classification, leaving other types of equally interesting learning tasks largely unexplored. A similar trend can be seen at the software level, as most published tools are designed for single-target classification and regression, and only a limited number of them focus on other types of learning tasks. One of those less explored areas involves the simultaneous prediction of multiple targets. Despite the broad applicability potential of the area of multi-target prediction (MTP), only a few tools have been proposed for specific subareas of multi-label classification. A more detailed review of such tools will be given in Section~\ref{sec:sec4}. 

The possibility of utilizing an automated tool for the majority of sub-areas that fall under the umbrella of MTP is an exciting idea. Typical examples of MTP settings are multi-label classification, multivariate regression, multi-task learning, dyadic prediction, zero-shot learning, network inference, and matrix completion. A first attempt in this direction was introduced by the DeepMTP framework~\cite{iliadis2022multi}, which will be reviewed in Section~2. This framework makes it possible for non-expert users to automatically select the most appropriate MTP problem setting by answering a handful of questions. After selecting the most appropriate MTP problem setting, the DeepMTP framework utilizes a flexible two-branch architecture that can be adjusted for specific MTP settings. The experiments presented in~\cite{iliadis2022multi} showcased DeepMTP as a competitive approach, compared to other baseline methods across multiple MTP problem settings and datasets. In terms of HPO, a standard grid search was used for all the comparisons. Even though this is acceptable in a research environment, it is certainly not practical for a user-centered software package. This is the primary motivation behind the work presented in this paper. We intend to increase the practical usability of the DeepMTP framework with an extension that utilizes efficient HPO methods. These HPO methods will be described in Section~3, and the benchmarking results will be presented in Section~5.

\section{A short review of DeepMTP}
\label{sec:sec2}
\subsection{Automated selection of the most suitable MTP problem setting}

\begin{figure}[t!]
\centering
  \includegraphics[trim=260 150 260 80,clip,width=\linewidth]{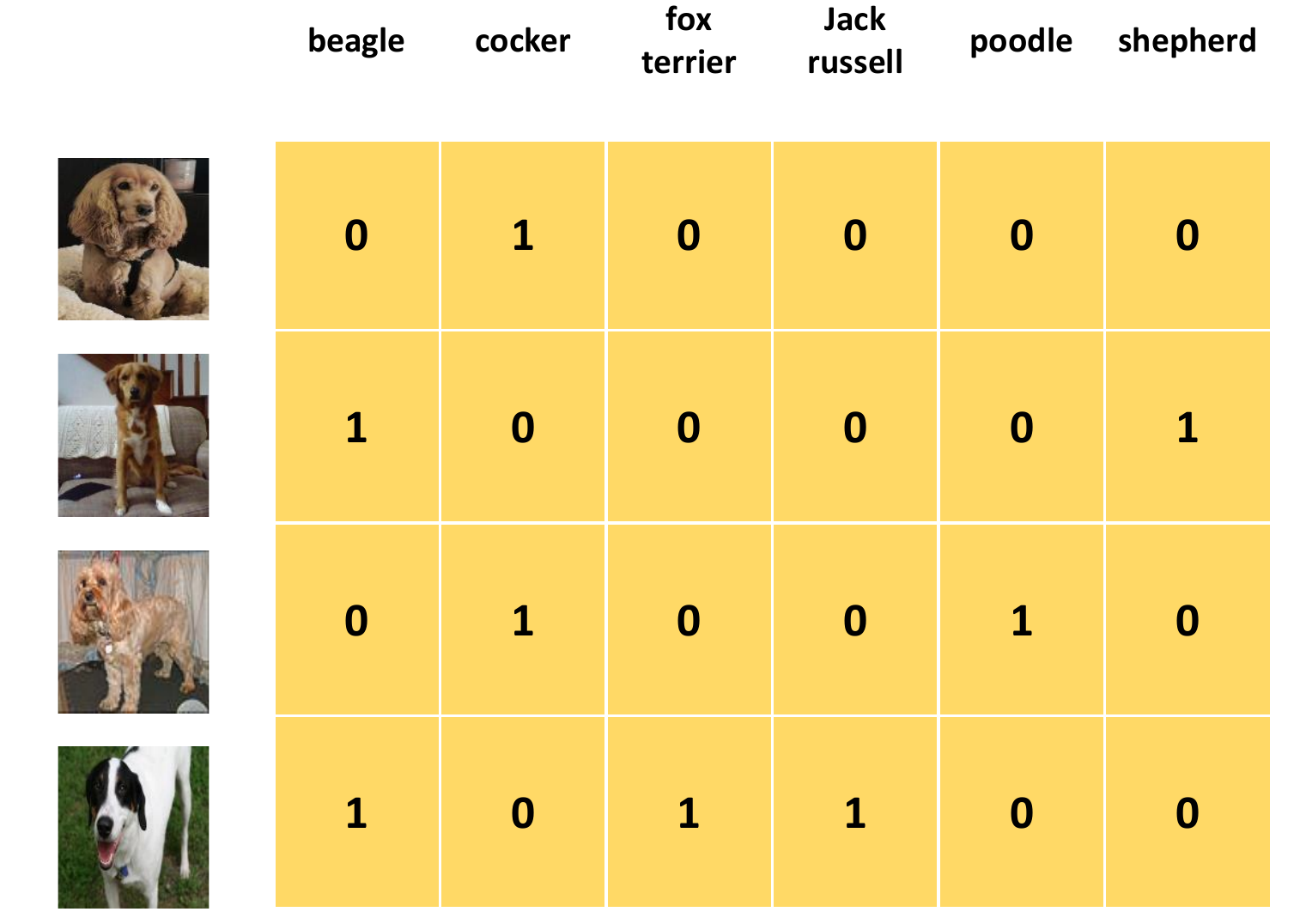}
\caption{Example of a multi-label classification task in which the goal is to identify the breeds of mixed-dogs. The instances correspond to images of dogs and the labels/targets to the breed names.}
\label{fig:MLC_example}
\end{figure}

As mentioned in the introduction, multi-target prediction comprises various sub-areas of machine learning, such as multi-label classification, multivariate regression, multi-task learning, dyadic prediction, zero-shot learning, network inference, and matrix completion. All these problem settings display specific characteristics that have resulted in the development of specific machine learning methods. At the same time, they also share a significant commonality, i.e., the prediction of multiple target variables. An example of a multi-label classification problem, the detection of dog breeds from images of mixed-breed dogs, can be seen in Fig.~\ref{fig:MLC_example}. The main difference in this task is that every dog can be associated with multiple breeds simultaneously (multi-label) instead of just one of them (single-label). Since a detailed formal definition of every MTP problem setting is out of scope for this work, we only present the general definition of MTP. For more details about the other MTP problem settings, we refer the reader to the survey by Waegeman et al.~\cite{waegeman2019multi}. A software package for DeepMTP is already available online\footnote{\url{https://github.com/diliadis/DeepMTP}} and a manuscript with a detailed explanation of the capabilities and general functionality is under review.

\begin{definition}\label{def:MTP_definition}
A \textbf{multi-target prediction problem} is characterized by an instance set $\mathcal{X}$,  a target set $\mathcal{T}$ and a score set $\mathcal{Y}$ with the following properties:
\begin{itemize}
\label{definition:MTP_definition}
\itemindent=10pt   
  \item [(P1)] A training dataset $\mathcal{D}$ contains triplets $(\mathbf{x}_i,\mathbf{t}_j,y_{ij})$,
  where $\mathbf{x}_i \in \mathcal{X}$ represents an instance, $\mathbf{t}_j \in \mathcal{T}$ represents a target, 
  and $y_{ij} \in \mathcal{Y}$ is the score that quantifies the relationship between an instance and a target, with $i\in\{1,\ldots,n\}$ and $j\in\{1,\ldots,m\}$. The scores can be arranged in an $n \times m$ matrix $\mathbf{Y}$
  that is usually incomplete.
       \item [(P2)] The score set $\mathcal{Y}$ consists of nominal, ordinal or real values.  
    \item [(P3)] During testing, the objective is to predict the score for any unobserved instance-target couple $(\mathbf{x},\mathbf{t}) \in \mathcal{X} \times \mathcal{T}$.
\end{itemize}
\end{definition}

This definition is very general, as it intends to cover all the MTP  problem settings that were considered by Waegeman et al.~\cite{waegeman2019multi}.
It describes three basic properties, so every MTP problem setting can be defined by adding more custom properties to the primary three. This is clear in the formal definition of multi-label classification, which requires four additional properties. Property~\textbf{P4} defines the generalization objective since one expects to make only predictions for new targets during testing. Property~\textbf{P5} specifies the absence of side information (commonly known as input features) for the targets. Finally, properties~\textbf{P6} and~\textbf{P7} inform about the state of the score matrix, mainly that the interaction values for all possible (instance, target) pairs $(\mathbf{x}_i,\mathbf{t}_j)$ in our training set are known and of binary type. These properties form the basis for the questionnaire used by the DeepMTP framework.

\begin{itemize}
\itemindent=8pt    \item [\textbf{Q1}:] Is it expected to encounter novel instances during testing? \textbf{(yes/no)}
\itemindent=8pt    \item [\textbf{Q2}:] Is it expected to encounter novel targets during testing? \textbf{(yes/no)}
\itemindent=8pt    \item [\textbf{Q3}:] Is there side information available for the instances? \textbf{(yes/no)}
\itemindent=8pt    \item [\textbf{Q4}:] Is there side information available for the targets? \textbf{(yes/no)}
\itemindent=8pt    \item [\textbf{Q5}:] Is the score matrix fully observed? \textbf{(yes/no)}
\itemindent=8pt    \item [\textbf{Q6}:] What is the type of the target variable? \textbf{(binary/nominal/ordinal/real-valued)}
\end{itemize}

Question \textbf{Q2} can be mapped to property \textbf{P4} as it relates to generalizing to new targets. Question \textbf{Q4} and property \textbf{P5} refer to the existence of side information for the targets. Questions \textbf{Q1} and \textbf{Q2} were generated from similar properties \textbf{P4} and \textbf{P5} that refer to the instances. Finally, questions \textbf{Q5} and \textbf{Q6} are derived from properties \textbf{P6} and \textbf{P7}, respectively (state of score matrix and target variable type).

Table~\ref{table:questionnarie_table} shows how specific combinations of answers lead to individual MTP problem settings. The mapping is not based on a data-driven method, but the in-house expertise from the research team that developed DeepMTP~\cite{waegeman2019multi, iliadis2022multi}. Furthermore, the selection of a specific MTP problem setting does not affect the configuration of the neural network architecture used by the DeepMTP framework, but intends to guide the user to the most appropriate literature. Configuration-related decisions are made based on the respective questions that comprise the questionnaire. The practical aspect of the framework moves away from the individual MTP settings to a more general view that is based on three principles. These include the generalization objectives for instances and targets, the existence of side information for instances and targets, and finally, the target variable type.

The questionnaire can be answered manually by (inexperienced) users or even automatically if the dataset is provided. There are at most three possible datasets that can be required by any of the MTP problem settings. Two of them contain the side information for the instances and targets, and the third one contains the score matrix, a prerequisite for every MTP problem setting. If these datasets are supplied to the framework, the task of answering the questionnaire becomes trivial with a simple computer program. If a user uploads the side information files for the instances and targets, questions \textbf{Q3} and \textbf{Q4} are answered trivially. Question~\textbf{Q5} can be answered by detecting missing values in the supplied dataset. To automatically determine the answers for questions \textbf{Q1} and \textbf{Q2}, one can compare the relations between instances (targets) in the training and test files, thus arriving at one of four generalization (validation) settings. Setting A involves the prediction of missing values inside the interaction matrix. In Setting B, the goal is to make predictions for new instances, while Setting C involves the prediction for new targets. Setting D requires the prediction for novel pairs of instances and targets. In conclusion, we show that in the first stage of the DeepMTP framework, the selection of the most appropriate MTP problem setting can be automated by combining the original questionnaire with basic programming.

\begin{table}[t]
\centering
    \caption{A snapshot of specific answers to the DeepMTP questionnaire and the corresponding MTP problem setting. ``-" denotes a wildcard. }
    \label{table:questionnarie_table}
    \begin{tabular}{c|c|c|c|c|c|c}
    \textbf{Q1} & \textbf{Q2} & \textbf{Q3} & \textbf{Q4} & \textbf{Q5} & \textbf{Q6} & \textbf{MTP method}        \\
    \hline
    yes         & no          & yes         & no          & yes         & binary      & Multi-label classification \\
    yes         & no          & yes         & no          & yes         & real-valued & Multivariate regression\\
    yes         & no          & yes         & no          & no          & -           & Multi-task learning        \\
    yes         & no          & yes         & yes (hierarchy)   & yes         & binary      & Hierarchical multi-label classification\\
    yes         & no          & yes         & yes         & no          & -           & Dyadic prediction\\
    yes         & yes         & yes         & yes         & no           & -           & Zero-shot learning\\
    no          & no          & no          & no          & no          & -           & Matrix completion\\
    no         & no         & yes           & yes           & no          & -           & Hybrid matrix completion\\
    yes        & yes         & yes          & yes           & no        &  -           & Cold-start collaborative filtering\\
    yes         & no          & yes         & no          & yes         & nominal/categorical           & Multi-dimensional classification~
    
    \end{tabular}
\end{table}

\subsection{The neural network architecture behind DeepMTP}

\begin{figure}[t!]
  \includegraphics[trim=75 155 75 120,clip,width=\linewidth]{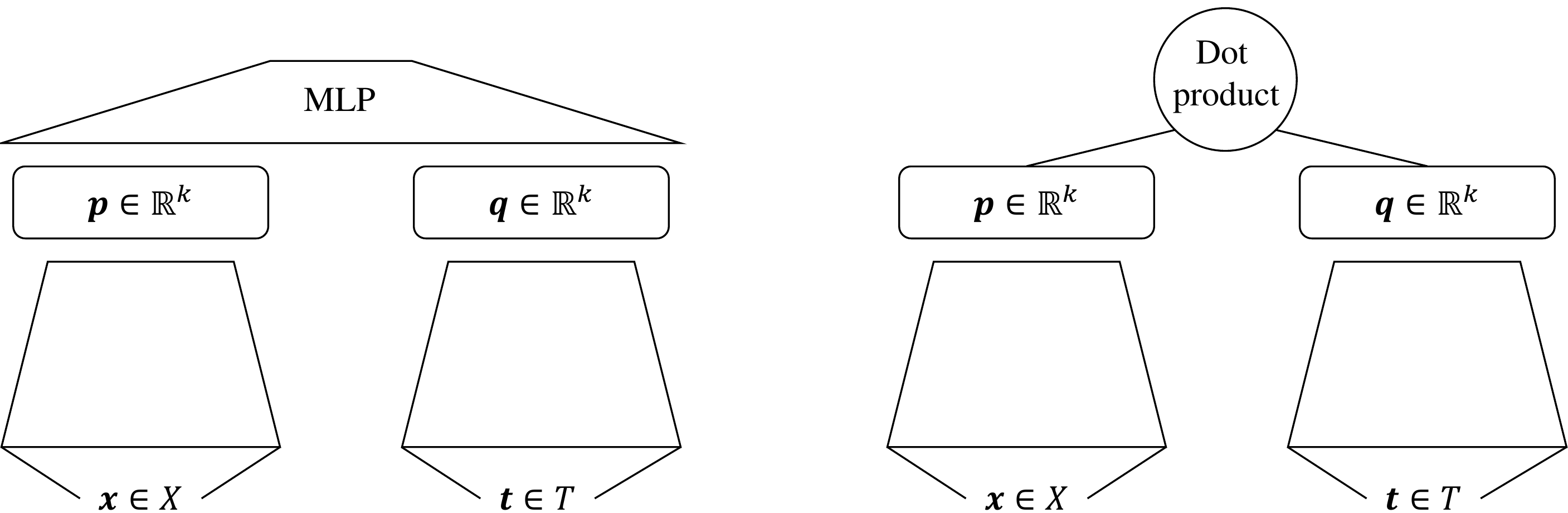}
\caption{Simplified view of the two two-branch neural network architectures.}
\label{fig:mlp_vs_dot}
\end{figure}

The DeepMTP framework utilizes a two-branch architecture that has gained popularity in the field of collaborative filtering~\cite{he2017neural}. However, the same architecture can be easily modified to achieve competitive performance across multiple MTP problem settings. The architecture (see Fig.~\ref{fig:mlp_vs_dot} left) features two branches that take as input any available side information for the instances and targets and then output two embedding vectors $\mathbf{p_x}$ and $\mathbf{q_t}$, respectively. The embeddings are then concatenated and the resulting vector is used as input to a series of fully connected layers that terminate at a single output node, as follows:
\begin{equation} \label{eq1}
\begin{split}
\mathbf{z}_1 & = \phi_1(\mathbf{p_x}, \mathbf{q_t}) = \left[\begin{array}{c}\mathbf{p_x} \\\mathbf{q_t} \end{array}\right]\,, \\
\phi_2(\mathbf{z}_1) & = \alpha_{2}(\mathbf{W}_{2}^T \mathbf{z}_1 + \mathbf{b}_2)\,, \\
& \vdots\\
\phi_L(\mathbf{z}_{L-1}) & = \alpha_{L}(\mathbf{W}_{L}^T \mathbf{z}_{L-1} + \mathbf{b}_L)\,, \\
\hat{y}_\mathbf{xt} & = \sigma(\mathbf{h}^T \phi_L (\mathbf{z}_{L-1}))\,,\\
\end{split}
\end{equation}
 where $\mathbf{W}$, $\mathbf{b}$ and $\alpha$ represent the weight matrix, bias vector, and activation function of the final multi-layer perceptron (MLP) layer, respectively. Alternatively, a seemingly more straightforward yet also less expressive approach skips the final series of fully-connected layers and instead computes the dot product (see Fig.~\ref{fig:mlp_vs_dot} right) of the two embedding vectors in the following way:
\begin{equation}
\hat{y}_\mathbf{xt} = \sigma(\mathbf{p_x} \cdot \mathbf{q_t})\,.
\end{equation}

Even though the architecture that uses the MLP was initially proposed as a more powerful approximator~\cite{he2017neural}, subsequent work~\cite{rendle2020neural, dacrema2021troubling} has argued that the reality of training the more complex model results in practical disadvantages. Our preliminary experiments seem to agree with that observation,
so all the experiments presented below will use the dot product version. 
Any modifications to this architecture are guided by the answers to the aforementioned questionnaire in the following way:
\begin{itemize}
    \item The combination of answers for \textbf{Q1} and \textbf{Q2} determines the validation setting the user expects. This can be explicitly requested by the user, especially if more than one validation setting is available given the datasets, or inferred by the relation of the instance and target IDs in the train and test datasets.
    \item The answers to questions \textbf{Q3} and \textbf{Q4} play multiple roles. A negative answer for any of the two questions will lead to the use of one-hot encoded vectors as the input to the corresponding branch. Furthermore, these answers determine the feasibility of the generalization of the user requests (through \textbf{Q1} and~\textbf{Q2}).
    \item Question \textbf{Q5} is used to distinguish between MTP problem settings (e.g., multi-label classification and multi-task learning).
    \item The answer to question \textbf{Q6} dictates the type of loss function that is used during training (binary cross-entropy loss for classification tasks and squared error loss for regression tasks).
\end{itemize}

To conclude, the neural network used in the DeepMTP framework is capable of adapting to the various characteristics that MTP problem settings exhibit,
from the existence of input features to the type of task that they represent. This functionality, combined with the purpose-made questionnaire and a basic user interface, can make multi-target prediction a more accessible area of research.

\section{Hyperparameter optimization methods}
\label{sec:sec3}
In this section, we give an overview of the hyperparameter optimization (HPO) methods that we consider for the benchmarking experiments in Section~\ref{sec:sec5}.
These techniques are usually grouped into two main categories: black box and multi-fidelity techniques.\\

\noindent \textbf{Grid Search \& Random Search:}
Conceptually, grid search is generally considered the simplest HPO method, as the optimal configuration is identified by brute-forcing the evaluation of all possible configurations of a user-defined hyperparameter space~\cite{montgomery2017design}.
The main weakness of this approach is its high computational cost, as the curse of dimensionality means that the number of configuration evaluations grows exponentially with the size of the configuration space. Furthermore, because continuous hyperparameters need to be discretized, an increase in their resolution can quickly lead to an explosion in the resulting number of configuration evaluations.

Instead of exhaustively searching over the hyperparameter space, a random search samples configurations at random until a user-defined budget is exhausted~\cite{bergstra2012random}. This approach is usually able to outperform grid search in cases where some hyperparameters are more important than others, as it may find configurations that are left out when discretizing. In the literature, random search is a standard baseline when comparing HPO methods, which is also why we choose to include a random search in the experimental section of this paper.\\

\noindent \textbf{Sequential Model-based Algorithm Configuration:}
The sequential model-based algorithm configuration (SMAC) approach~\cite{hutter2011sequential} is a Bayesian optimization method that uses random forests as a surrogate model. The main advantage over other options such as Gaussian processes is that random forests can naturally support categorical hyperparameters, handle larger search spaces, and scale better as the number of training samples increases.

After an initialization phase of randomly sampled observations, SMAC fits a random forest to the collected observations, serving as the surrogate model in the framework of Bayesian optimization. Subsequently, SMAC alternates between determining the next hyperparameter configuration to evaluate and updating the surrogate model with the newly evaluated observation. In the former step, the surrogate model is employed to estimate the usefulness of evaluating a hyperparameter configuration $\theta$. Assessing the usefulness of an evaluation requires tackling the so-called exploration-exploitation dilemma, which is traditionally implemented via a so-called acquisition function, e.g., the expected improvement $\mathbb{EI}(\theta)$, comparing the potential improvement of some configuration over the best configuration observed so far:
\begin{equation}
\mathbb{EI}(\theta) = \sigma_{\theta}[u \cdot \Phi(u)+\phi(u)],  \quad  u = \frac{o_{min}-\mu_{\theta}}{\sigma_{\theta}}
\end{equation}
where $\Phi$ denotes the cumulative distribution function of the standard normal distribution,
$\phi$ is the corresponding probability density function, and $o_{min}$ is the loss of the best performing configuration. This approach has proven to be quite successful and is at the heart of several AutoML software packages (e.g., Auto-Weka~\cite{thornton2013auto}, auto-sklearn~\cite{feurer-neurips15a}). For our experiments, we used the SMAC3 implementation~\cite{lindauer2021smac3} that is available online\footnotemark.\\
\footnotetext{\url{https://github.com/automl/SMAC3}}

\noindent \textbf{Multi-fidelity optimization - Hyperband: } One of the core steps in any standard HPO method is the performance evaluation of a given configuration. This can be manageable for simple models that are relatively cheap to train and test, but can become a significant bottleneck for more complex models that need hours or even days to train. This is particularly evident in deep learning, as big neural networks with millions of parameters trained on increasingly larger datasets can deem traditional black-box HPO methods impractical.

Addressing this issue, multi-fidelity HPO methods have been devised to discard unpromising hyperparameter configurations already at an early stage. To this end, the evaluation procedure is adapted to support cheaper evaluations of hyperparameter configurations, such as evaluating on sub-samples (feature-wise or instance-wise) of the provided data set or executing the training procedure only for a certain number of epochs in the case of iterative learners. The more promising candidates are subsequently evaluated on increasing budgets until a maximum assignable budget is reached.

A popular representative of such methods is Hyperband~\cite{DBLP:journals/jmlr/LiJDRT17}. Hyperband builds upon Successive Halving (SH)~\cite{DBLP:conf/icml/KarninKS13}, where a set of $n$ candidates is first evaluated on a small budget. Based on these \textit{low-fidelity} performance estimates, the $\frac{n}{\eta}$ ($\eta \geq 2)$ best candidates are preserved, while the remaining configurations are already discarded. Iteratively increasing the evaluation budget and re-evaluating the remaining candidates with the increased budget while discarding the inferior candidates results in fewer resources wasted on inferior candidates. In return, one focuses more on the promising candidates.

Despite the efficiency of the successive halving strategy, it is well known that it suffers from the exploration-exploitation trade-off. In simple terms, a static budget $\mathcal{B}$ means that the user has to manually decide whether to explore a number of configurations $n$ or give each configuration a sufficient budget to develop. An incorrect decision can lead to an inadequate exploration of the search space (small $n$) or the early rejection of promising configurations (large $n$). Hyperband overcomes the exploration-exploitation trade-off by repeating the successive halving strategy with different initializations of SH, varying the budget and the number of initial candidate configurations.

\noindent \textbf{Bayesian Optimization with Hyperband: }Despite the advantages that Hyperband displays compared to baseline methods like grid search and random search, it is still restricted by the random sampling of configurations at the beginning of each iteration of the successive halving routine. Learning from past sampled configurations has the potential to provide improvements in the final performance compared to standard, model-free Hyperband. The Bayesian optimization with Hyperband method~\cite{falkner2018bohb} tries to improve over this by replacing random sampling with a model-based approach that utilizes Bayesian optimization.

More specifically, at the beginning of each bracket, Hyperband determines the number of configurations, and the Bayesian optimization component decides which configurations to consider. 
While initially the latter will suggest configurations randomly, once sufficiently many observations have been made, a surrogate model is fitted to those observations, and it is used to suggest new hyperparameter configurations that maximize the expected improvement acquisition function. The experiments presented by Falkner et al.~\cite{falkner2018bohb} support that this model-based approach outperforms other model-free baselines such as random search and grid search.

\section{Related Work}
\label{sec:sec4}
Recent advances in the theoretical and methodological aspects of AutoML have been closely followed by the publication of accompanying software. These open-source frameworks work as a test bench of AutoML theory in the real world. A software package called Auto-WEKA~\cite{thornton2013auto} was one of the first attempts to offer an implementation to tackle the combined algorithm selection and hyperparameter optimization (CASH) problem. The package optimizes over the classification models and feature selectors offered by the original WEKA package using the SMAC optimizer detailed above.
Hyperopt-Sklearn~\cite{komer2014hyperopt} is another project designed for the CASH problem. The optimization component uses the Hyperopt library, another implementation of Bayesian optimization, and the baseline models are provided by the popular scikit-learn library. Hyperopt is designed to optimize over the search domain that is generated by the combinations of scikit-learn's preprocessing, classification and regression modules. Similar to Auto-WEKA, the search domain can be comprised of random variables that are sampled and then mapped by an objective function to a scalar score. The score can then be minimized by any of the supported optimizers (Random Search, TPE, Gaussian Process Trees). Auto-sklearn~\cite{Feurer2019} is another AutoML system that uses scikit-learn's implemented preprocessing, classification, and regression modules to define the configuration space. In this implementation, SMAC is used for optimizing over a hypothesis space. The model-based optimization approach was designed to use performance data from similar datasets and to construct ensembles of baseline models evaluated during the optimization state.

All of the tools mentioned above are designed for single-target classification and regression problems. Generalizing to multiple targets, several frameworks have been published and gained attention in their respective fields. MULAN is a popular Java library~\cite{tsoumakas2011mulan} built on top of the well-known WEKA platform and provides a wide range of multi-label classification and multivariate regression algorithms. Another open-source library called scikit-multilearn~\cite{2017arXiv170201460S} was published more recently, providing fewer methods compared to MULAN, but it is written using the more popular Python language. In the area of hierarchical multi-label classification, Clus~\cite{vens2008decision} is a decision tree-based open-source framework that provides a Java interface. Despite the popularity of these tools in specific sub-areas of MTP, none of them offer any automation options in the algorithm selection or hyperparameter optimization steps. In the area of multi-label classification, the work of de Sa et al.~\cite{de2017towards, de2018automated} uses genetic algorithms in the first attempt at automating the task. Furthermore, Wever et al. proposed an extension of ML-Plan~\cite{wever2019automating}, an approach that combines hierarchical task network planning with a best-first search, to configure multi-label classifiers and managed to outperform other baselines in multi-label classification benchmarks, including the ones proposed by de Sa et al.~\cite{de2017towards, de2018automated}. Finally, CascadeML~\cite{pakrashi2019cascademl} proposed a cascade neural network that utilizes label associations and requires minimal hyperparameter tuning as another viable benchmark for multi-label classification datasets.

In recent years, machine learning frameworks like Pytorch~\cite{paszke2019pytorch} and Tensorflow~\cite{abadi2016tensorflow} have gained considerable popularity in the area of deep learning. As a result, AutoML libraries suited specifically for these deep learning frameworks are now being released. Auto-Net is one of the first attempts at automatically tuning neural network architectures. Its first major release, Auto-Net 1.0~\cite{mendoza2016towards}, uses the SMAC optimizer and Lasagne~\cite{sander_dieleman_2015_27878} 
as the deep learning framework, a seemingly powerful combination, as it was one of the first to outperform expert users on competition datasets~\cite{guyon2015design}. Auto-Net 2.0, first described in a book chapter~\cite{mendoza2019towards}, was able to bring performance improvements over the first release by replacing SMAC with BOHB. Those ideas were further improved in~\cite{zimmer2021auto} with the introduction of Auto-Pytorch, a framework that combines ensembling with multi-fidelity optimization and meta-learning and, as a result, brings significant efficiencies.

To conclude, the aforementioned methods and software packages show that multi-target prediction is a largely unexplored area, which has seen some recent interest in one of its most popular problem settings. In this work, we attempt to set the beginning stages of a similar software package specifically adapted for all the problem settings that fall under the umbrella of MTP. This is achieved by using the automatically answered questionnaire to select the most appropriate MTP setting, and then deploy one of the HPO methods that we benchmark in the next section, on a flexible two-branch neural network architecture.

\section{Evaluation of HPO methods for DeepMTP}
\label{sec:sec5}

\begin{table}[t!]
\centering
    \caption{Basic information about the datasets using in the experiments across five MTP problem settings.}
    \label{table:dataset_stats_table}
\resizebox{\textwidth}{!}{
\begin{tabular}{lc|cccc}
                                                     & \multicolumn{1}{l|}{} & \# instances & \# targets & \# instance features & \# target features \\ \hline
\multirow{2}{*}{\textbf{Multi-label classification}} & Bibtex                & 7395         & 159        & 1836                 & 159                \\
                                                     & Corel5k               & 5000         & 374        & 499                  & 374                \\ \hline
\multirow{2}{*}{\textbf{Multivariate regression}}    & Rf2                   & 9125         & 8          & 576                  & 8                  \\
                                                     & scm1d                 & 9803         & 16         & 280                  & 16                 \\ \hline
\multirow{2}{*}{\textbf{Multi-task learning}}        & dog                   & 800          & 52         & 3*224*224            & 52                 \\
                                                     & bird                  & 2000         & 65         & 3*224*224            & 65                 \\ \hline
\multirow{2}{*}{\textbf{Matrix completion}}          & Movielens100k         & 943          & 1682       & 943                  & 1682               \\
                                                     & Movielens1M           & 6040         & 3706       & 6040                 & 3706               \\ \hline
\multirow{2}{*}{\textbf{Dyadic prediction}}          & ERN                   & 1164         & 154        & 445                  & 445                \\
                                                     & SRN                   & 1821         & 9884       & 113                  & 1685               \\ \hline
\end{tabular}}
\end{table}

This section's goal is to compare Random Search, Random Search with a doubled budget, Hyperband, BOHB, and SMAC3 across different MTP settings, task types (classification, regression), dataset sizes, and types.

\subsection{Experiment setup}
Basic information about all the datasets used for benchmarking is available in Table~\ref{table:dataset_stats_table}. Every dataset is split into training and test sets (80-20\%).  We also randomly sample 20\% from the training dataset to form an internal validation set that we use for early stopping and to determine the best configurations while optimizing the network. Also, every HPO run is repeated 5 times, and we report the average performance. The maximum budget allowed for the HPO methods like Hyperband, BOHB, and SMAC is different for every dataset. This parameter is determined by calculating the average best epoch from 20 randomly selected configurations tested before the HPO methods are benchmarked. For Hyperband and BOHB, every other parameter is set to the default values. The configuration space is similar for most of the experiments, with some additional restrictions introduced in cases where a branch encodes one-hot encoded vectors (only one layer allowed).
A primary goal of this work is to identify potential cases where one of the HPO methods could provide a clear advantage. That information can then be used to determine the HPO method suggested by the DeepMTP framework, further minimizing the number of inputs a regular user has to provide. Space constraints do not allow us to present detailed plots for every MTP problem setting and dataset, so in this section, we showcase a limited, yet representative number of examples. Detailed information about the hyperparameter spaces used for every dataset and all additional results and extensive visualizations are publicly available in a web-based application\footnotemark.

\footnotetext{\url{https://share.streamlit.io/anonymousmlresearcher/deepmtp_hpo_results/main/streamlit_interface.py}}

\begin{figure}[t!]
  \includegraphics[trim=1 58 3 63,clip,width=\linewidth]{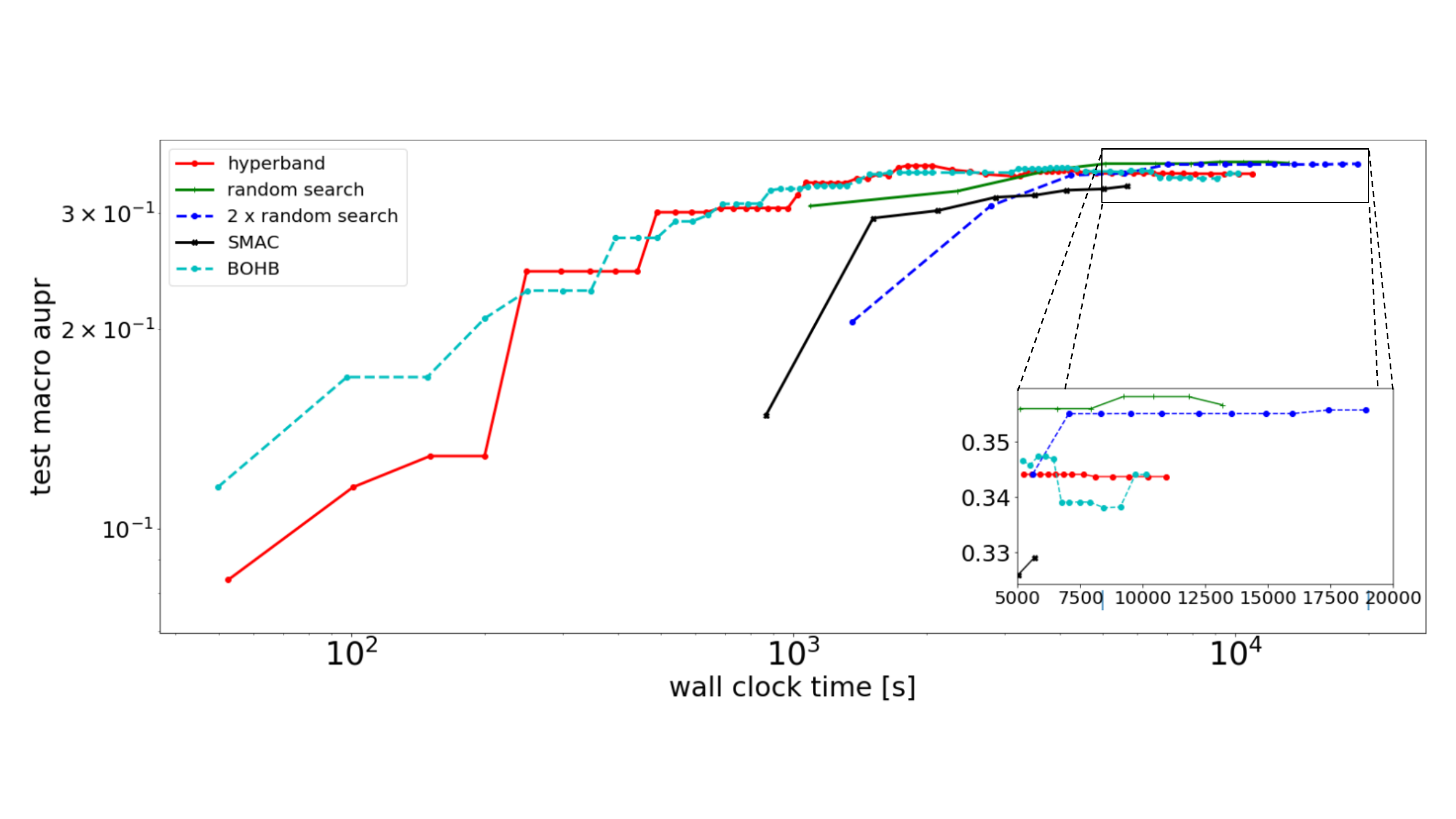}

  \includegraphics[trim=1 45 3 60,clip,width=\linewidth]{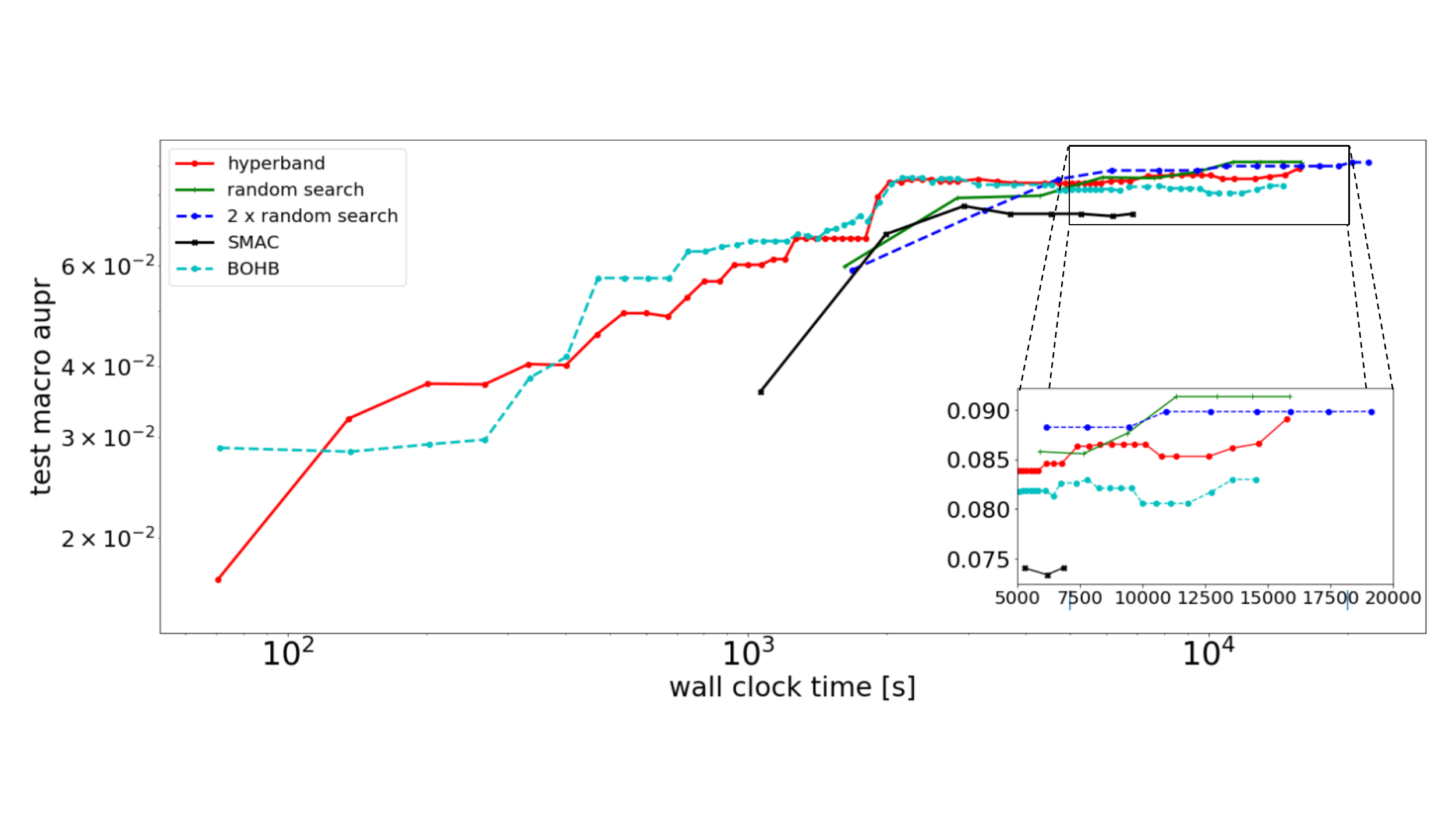}
\caption{Average test macro-AUPR of every HPO method across time for Bibtex (top) and Corel5k (bottom)}
\label{fig:mlc}
\end{figure}

\subsection{Comparing performance across time}

For the multi-label classification problem setting, we selected two of the largest datasets available in the Mulan repository~\cite{tsoumakas2011mulan}. The maximum budget allocated for both datasets was set to 27 epochs. In Fig.~\ref{fig:mlc}, we observe that random search is quite competitive with Hyperband and BOHB, while SMAC shows the worst performance out of the five candidates. Hyperband provides an early speed-up on the bibtex dataset (2.2 times compared to random search and 5.7 times compared to random search with double the budget) as well as corel5k dataset (1.8 times compared to random search and 2.3 times compared to random search with double the budget). Another interesting fact is that BOHB fails to outperform the standard Hyperband approach, something that might be attributed to the relatively small budget.

\begin{figure}[t!]
  \includegraphics[trim=0 0 0 0,,clip,width=\linewidth]{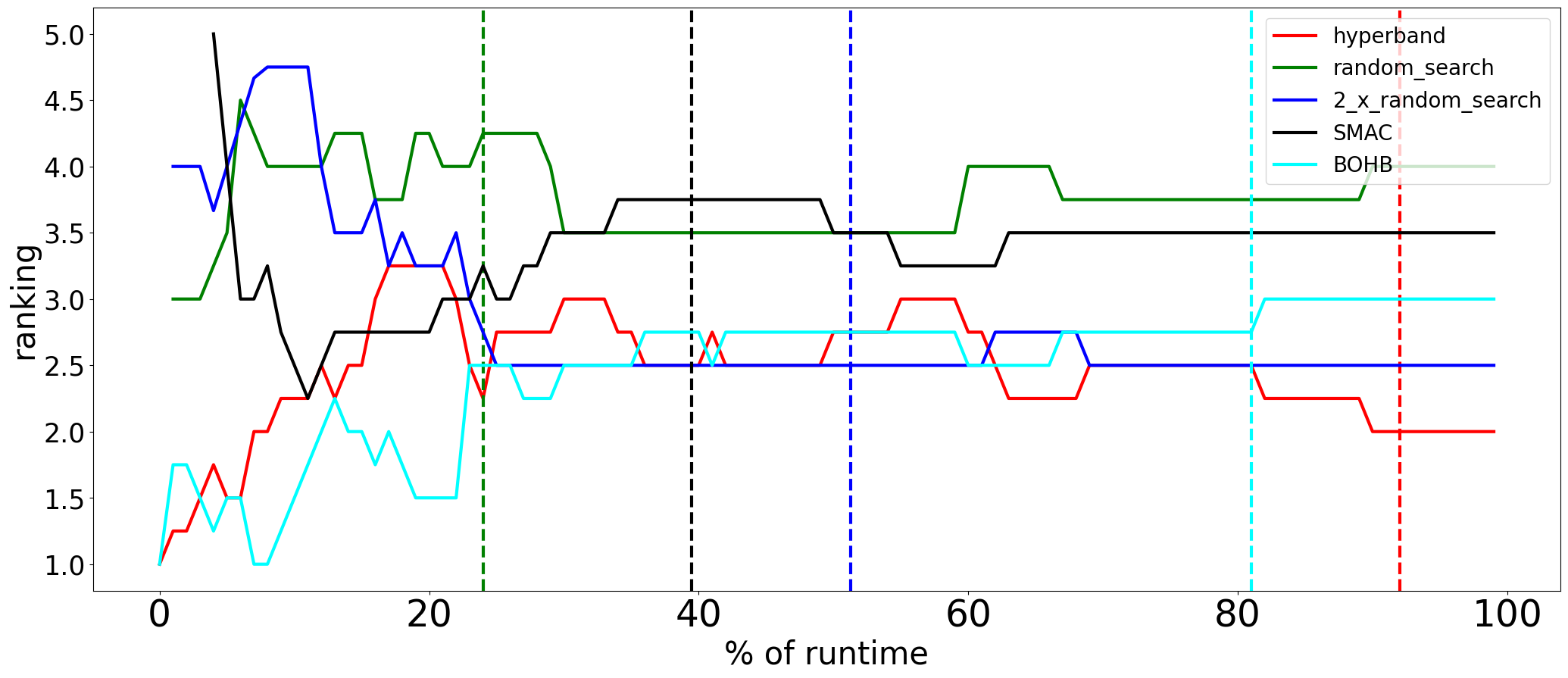}
  \includegraphics[trim=0 0 0 0,,clip,width=\linewidth]{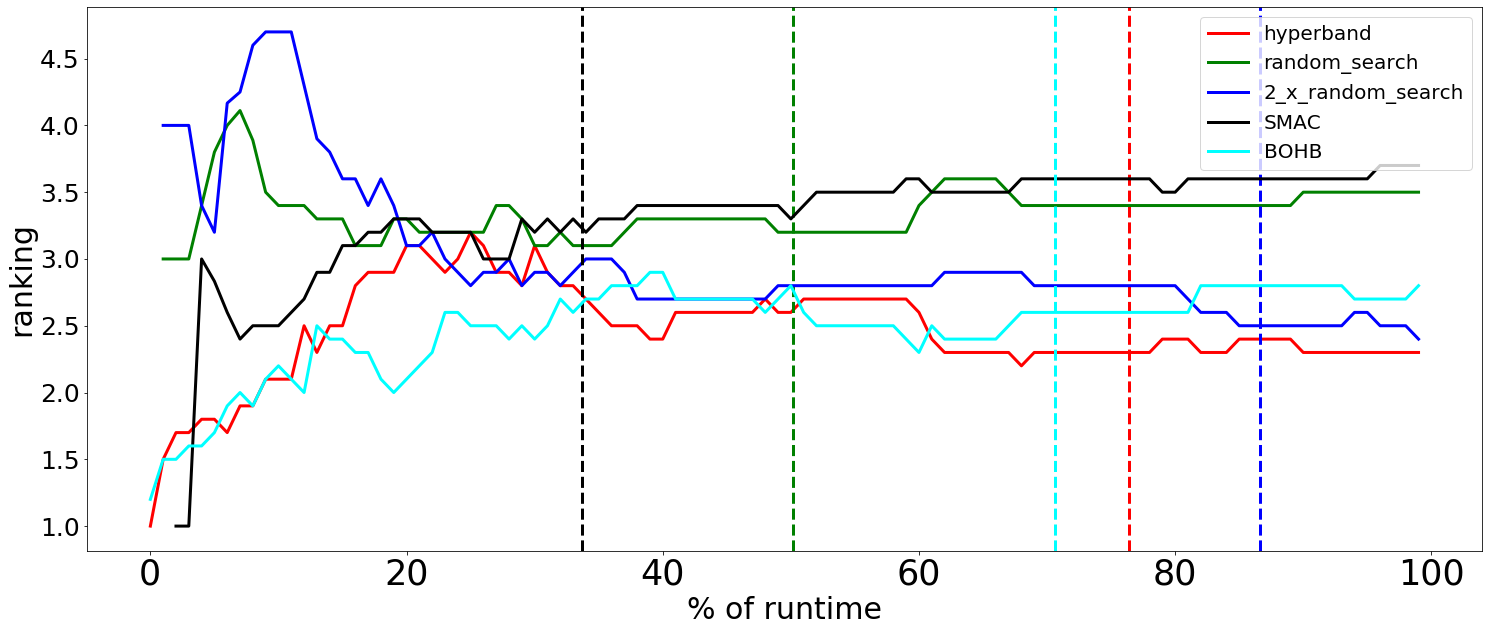}
\caption{Average ranking of every HPO method across the percentage of the respective runtimes for the MTP setting. In the y axis the lower values indicate higher ranking (lower is better). The vertical dotted lines represent the average end-points of every HPO approach. In terms of the performance metrics used to calculate the metrics, we decided to use the most frequently used metrics and averaging schemes for every MTP problem setting (macro-AUPR for the multi-label classification and multi-task learning datasets, micro-AUPR for the dyadic prediction datasets, macro-RRMSE for the multivariate regression datasets, and micro-RMSE for the matrix completion datasets). The top panel visualizes the average ranking over the regression datasets and the bottom panel the global ranking over all datasets.}
\label{fig:global_ranking}
\end{figure}

In order to generalize across MTP problem settings and provide a more informative comparison between the HPO methods considered, we generated a global ranking plot, shown in the bottom panel of Fig.~\ref{fig:global_ranking}. Based on this ranking, we observe that Hyperband and BOHB outperform the other three approaches for the first 30-40\% of the runtime. The same behavior is observed in the average ranking at the classification (multi-label classification, multi-task learning, and dyadic prediction) and regression level (multivariate regression and matrix completion), shown in the top panel of Fig.~\ref{fig:global_ranking}, with Hyperband and BOHB outranking the others. Diving deeper into the rankings per MTP problem setting, Hyperband ranks in the top 2 across all five settings in the early stages of runtime. After 30-40\% of the runtime has passed, random search shows a similar performance as Hyperband and BOHB, sometimes outperforming them. This effect can be seen in Fig.~\ref{fig:global_ranking}, as random search achieves a similar ranking as Hyperband at the very end of the total runtime. Because the ranking plots do not inform us about the actual performance difference between the best HPO methods, we decided to generate plots that quantify them. The results show that in the classification and  regression settings, the actual performance difference of the top-2 ranked HPO methods is relatively small. Space limitation does not allow us to present these plots here, so we make them available in the aforementioned web-based application\footnotemark[\value{footnote}].

Compared to Hyperband, the similar performance of the two more advanced Bayesian approaches, BOHB and SMAC, likely results from the relatively small budget we assign. Specifically for SMAC, the available budget ranges from eight to 10 configurations across all experiments, which is insufficient for the underlying optimizer. Even though SMAC and BOHB can show potential improvements by increasing the maximum budget, we argue that the standing comparison is fair and more representative of the time constraints that users impose in a real-world environment. The importance of getting relatively good results fairly quickly is the main advantage of Hyperband in the results we have obtained. If the user favors performance over runtime, a random search with a doubled budget can provide a similar or marginally better performance in some MTP problem settings.

\section{Conclusions}
\label{sec:sec6}

The goal of this paper was to present a fully-automated deep learning pipeline for multi-target prediction by extending the DeepMTP framework with hyperparameter optimization methods. We benchmarked the most popular HPO methods on ten different benchmark datasets from five different MTP settings. An important finding from our experiments is that BOHB and SMAC, two seemingly more advanced methods, do not result in any performance improvements compared to the standard Hyperband. Based on the results, we also conclude that Hyperband is a viable option for the majority of the MTP problem settings that our framework considers, as it provides significant speed-ups compared to the other baselines. Despite the competitiveness of random search, Hyperband is able to return results early in the optimization process, thus providing the option to stop the optimization if the performance is deemed adequate by the user.

\section*{Declarations}

\begin{itemize}[leftmargin=*,align=left]

\item[\textbf{Funding}]   This research received funding from the Flemish Government under the “Onderzoeksprogramma Artifici\"ele Intelligentie (AI) Vlaanderen” programme.

\item[\textbf{Conflicts of interest}]  The authors declare that they have no confict of interest.

\item[\textbf{Ethics approval}]  Not applicable.

\item[\textbf{Consent to participate}]  Not applicable.

\item[\textbf{Consent for publication}]  Not applicable.

\item[\textbf{Availability of data and material}]  The data used for the experiments are available online, see Section \ref{sec:sec6} for more details.

\item[\textbf{Code availability}] The code used to run the experiments can be found on github\footnotemark.
\item[\textbf{Authors' contributions}]  The first author implemented the python package. All four authors contributed equally to the manuscript.
\end{itemize}

\footnotetext{\url{https://github.com/diliadis/DeepMTP}}
%
%
%
\bibliographystyle{splncs04}
\bibliography{bibliography}

\begin{thebibliography}{10}
\providecommand{\url}[1]{\texttt{#1}}
\providecommand{\urlprefix}{URL }
\providecommand{\doi}[1]{https://doi.org/#1}

\bibitem{abadi2016tensorflow}
Abadi, M., Barham, P., Chen, J., Chen, Z., Davis, A., Dean, J., Devin, M.,
  Ghemawat, S., Irving, G., Isard, M., et~al.: {TensorFlow}: A system for
  {Large-Scale} machine learning. In: 12th USENIX symposium on OSDI 16. pp.
  265--283 (2016)

\bibitem{bergstra2012random}
Bergstra, J., Bengio, Y.: Random search for hyper-parameter optimization. J
  Mach Learn Res  \textbf{13}(2) (2012)

\bibitem{DBLP:journals/corr/abs-2107-05847}
Bischl, B., Binder, M., Lang, M., Pielok, T., Richter, J., Coors, S., Thomas,
  J., Ullmann, T., Becker, M., Boulesteix, A., Deng, D., Lindauer, M.:
  Hyperparameter optimization: Foundations, algorithms, best practices and open
  challenges. CoRR  \textbf{abs/2107.05847} (2021),
  \url{https://arxiv.org/abs/2107.05847}

\bibitem{dacrema2021troubling}
Dacrema, M.F., Boglio, S., Cremonesi, P., Jannach, D.: A troubling analysis of
  reproducibility and progress in recommender systems research. ACM TOIS
  \textbf{39}(2),  1--49 (2021)

\bibitem{sander_dieleman_2015_27878}
Dieleman, S., Schlüter, J., Raffel, C., Olson, E., Sønderby, S.K., Nouri, D.,
  Maturana, D., Thoma, M., Battenberg, E., Kelly, J., Fauw, J.D., Heilman, M.,
  diogo149, McFee, B., Weideman, H., takacsg84, peterderivaz, Jon, instagibbs,
  Rasul, D.K., CongLiu, Britefury, Degrave, J.: Lasagne: First release. (Aug
  2015). \doi{10.5281/zenodo.27878}, \url{https://doi.org/10.5281/zenodo.27878}

\bibitem{elsken2019neural}
Elsken, T., Metzen, J.H., Hutter, F.: Neural architecture search: A survey. J
  Mach Learn Res  \textbf{20}(1),  1997--2017 (2019)

\bibitem{falkner2018bohb}
Falkner, S., Klein, A., Hutter, F.: Bohb: Robust and efficient hyperparameter
  optimization at scale. In: ICML. pp. 1437--1446. PMLR (2018)

\bibitem{feurer-neurips15a}
Feurer, M., Klein, A., Eggensperger, K., Springenberg, J., Blum, M., Hutter,
  F.: Efficient and robust automated machine learning. In: NeurIPS. pp.
  2962--2970 (2015)

\bibitem{Feurer2019}
Feurer, M., Klein, A., Eggensperger, K., Springenberg, J.T., Blum, M., Hutter,
  F.: Auto-sklearn: Efficient and Robust Automated Machine Learning, pp.
  113--134. Springer International Publishing, Cham (2019)

\bibitem{goodfellow2014generative}
Goodfellow, I., Pouget-Abadie, J., Mirza, M., Xu, B., Warde-Farley, D., Ozair,
  S., Courville, A., Bengio, Y.: Generative adversarial nets. NeurIPS
  \textbf{27} (2014)

\bibitem{guyon2015design}
Guyon, I., Bennett, K., Cawley, G., Escalante, H.J., Escalera, S., Ho, T.K.,
  Maci{\`a}, N., Ray, B., Saeed, M., Statnikov, A., et~al.: Design of the 2015
  chalearn automl challenge. In: 2015 International Joint Conference on Neural
  Networks (IJCNN). pp.~1--8. IEEE (2015)

\bibitem{he2017neural}
He, X., Liao, L., Zhang, H., Nie, L., Hu, X., Chua, T.S.: Neural collaborative
  filtering. In: Proceedings of the 26th WWW. pp. 173--182 (2017)

\bibitem{he2021automl}
He, X., Zhao, K., Chu, X.: Automl: A survey of the state-of-the-art. Knowl.
  Based. Syst.  \textbf{212},  106622 (2021)

\bibitem{hutter2011sequential}
Hutter, F., Hoos, H.H., Leyton-Brown, K.: Sequential model-based optimization
  for general algorithm configuration. In: LION. pp. 507--523. Springer (2011)

\bibitem{DBLP:books/sp/HKV2019}
Hutter, F., Kotthoff, L., Vanschoren, J. (eds.): Automated Machine Learning -
  Methods, Systems, Challenges. The Springer Series on Challenges in Machine
  Learning, Springer (2019). \doi{10.1007/978-3-030-05318-5},
  \url{https://doi.org/10.1007/978-3-030-05318-5}

\bibitem{iliadis2022multi}
Iliadis, D., De~Baets, B., Waegeman, W.: Multi-target prediction for dummies
  using two-branch neural networks. Mach. Learn pp. 1--34 (2022)

\bibitem{jumper2021highly}
Jumper, J., Evans, R., Pritzel, A., Green, T., Figurnov, M., Ronneberger, O.,
  Tunyasuvunakool, K., Bates, R., {\v{Z}}{\'\i}dek, A., Potapenko, A., et~al.:
  Highly accurate protein structure prediction with alphafold. Nature
  \textbf{596}(7873),  583--589 (2021)

\bibitem{DBLP:conf/icml/KarninKS13}
Karnin, Z.S., Koren, T., Somekh, O.: Almost optimal exploration in multi-armed
  bandits. In: {ICML} 2013, Atlanta, GA, USA, 16-21 June 2013. JMLR, vol.~28,
  pp. 1238--1246. JMLR.org (2013),
  \url{http://proceedings.mlr.press/v28/karnin13.html}

\bibitem{komer2014hyperopt}
Komer, B., Bergstra, J., Eliasmith, C.: Hyperopt-sklearn: automatic
  hyperparameter configuration for scikit-learn. In: ICML workshop on AutoML.
  vol.~9, p.~50. Citeseer (2014)

\bibitem{krizhevsky2012imagenet}
Krizhevsky, A., Sutskever, I., Hinton, G.E.: Imagenet classification with deep
  convolutional neural networks. NeurIPS  \textbf{25} (2012)

\bibitem{DBLP:journals/jmlr/LiJDRT17}
Li, L., Jamieson, K.G., DeSalvo, G., Rostamizadeh, A., Talwalkar, A.:
  Hyperband: {A} novel bandit-based approach to hyperparameter optimization. J.
  Mach. Learn. Res.  \textbf{18},  185:1--185:52 (2017),
  \url{http://jmlr.org/papers/v18/16-558.html}

\bibitem{lindauer2021smac3}
Lindauer, M., Eggensperger, K., Feurer, M., Biedenkapp, A., Deng, D.,
  Benjamins, C., Ruhkopf, T., Sass, R., Hutter, F.: Smac3: A versatile bayesian
  optimization package for hyperparameter optimization (2021)

\bibitem{mendoza2016towards}
Mendoza, H., Klein, A., Feurer, M., Springenberg, J.T., Hutter, F.: Towards
  automatically-tuned neural networks. In: Workshop on Automatic Machine
  Learning. pp. 58--65. PMLR (2016)

\bibitem{mendoza2019towards}
Mendoza, H., Klein, A., Feurer, M., Springenberg, J.T., Urban, M., Burkart, M.,
  Dippel, M., Lindauer, M., Hutter, F.: Towards automatically-tuned deep neural
  networks. In: Automated machine learning, pp. 135--149. Springer, Cham (2019)

\bibitem{montgomery2017design}
Montgomery, D.C.: Design and analysis of experiments. John wiley \& sons (2017)

\bibitem{pakrashi2019cascademl}
Pakrashi, A., Mac~Namee, B.: Cascademl: An automatic neural network
  architecture evolution and training algorithm for multi-label classification
  (best technical paper). In: SGAI. pp. 3--17. Springer (2019)

\bibitem{paszke2019pytorch}
Paszke, A., Gross, S., Massa, F., Lerer, A., Bradbury, J., Chanan, G., Killeen,
  T., Lin, Z., Gimelshein, N., Antiga, L., et~al.: Pytorch: An imperative
  style, high-performance deep learning library. NeurIPS  \textbf{32} (2019)

\bibitem{pham2018efficient}
Pham, H., Guan, M., Zoph, B., Le, Q., Dean, J.: Efficient neural architecture
  search via parameters sharing. In: ICML. pp. 4095--4104. PMLR (2018)

\bibitem{rendle2020neural}
Rendle, S., Krichene, W., Zhang, L., Anderson, J.: Neural collaborative
  filtering vs. matrix factorization revisited. In: Fourteenth ACM Conference
  on RecSys. pp. 240--248 (2020)

\bibitem{de2018automated}
de~S{\'a}, A.G., Freitas, A.A., Pappa, G.L.: Automated selection and
  configuration of multi-label classification algorithms with grammar-based
  genetic programming. In: PPSN. pp. 308--320. Springer (2018)

\bibitem{de2017towards}
de~S{\'a}, A.G., Pappa, G.L., Freitas, A.A.: Towards a method for automatically
  selecting and configuring multi-label classification algorithms. In: GECCO.
  pp. 1125--1132 (2017)

\bibitem{simonyan2014very}
Simonyan, K., Zisserman, A.: Very deep convolutional networks for large-scale
  image recognition. arXiv preprint arXiv:1409.1556  (2014)

\bibitem{2017arXiv170201460S}
{Szyma{\'n}ski}, P., {Kajdanowicz}, T.: {A scikit-based Python environment for
  performing multi-label classification}. ArXiv e-prints  (Feb 2017)

\bibitem{thornton2013auto}
Thornton, C., Hutter, F., Hoos, H.H., Leyton-Brown, K.: Auto-weka: Combined
  selection and hyperparameter optimization of classification algorithms. In:
  Proceedings of the 19th ACM SIGKDD. pp. 847--855 (2013)

\bibitem{tsoumakas2011mulan}
Tsoumakas, G., Spyromitros-Xioufis, E., Vilcek, J., Vlahavas, I.: Mulan: A java
  library for multi-label learning. J Mach. Learn. Res.  \textbf{12},
  2411--2414 (2011)

\bibitem{vaswani2017attention}
Vaswani, A., Shazeer, N., Parmar, N., Uszkoreit, J., Jones, L., Gomez, A.N.,
  Kaiser, {\L}., Polosukhin, I.: Attention is all you need. NeurIPS
  \textbf{30} (2017)

\bibitem{vens2008decision}
Vens, C., Struyf, J., Schietgat, L., D{\v{z}}eroski, S., Blockeel, H.: Decision
  trees for hierarchical multi-label classification. Machine learning
  \textbf{73}(2),  185--214 (2008)

\bibitem{waegeman2019multi}
Waegeman, W., Dembczy{\'n}ski, K., H{\"u}llermeier, E.: Multi-target
  prediction: a unifying view on problems and methods. Data Min Knowl Discov
  \textbf{33}(2),  293--324 (2019)

\bibitem{wever2019automating}
Wever, M.D., Mohr, F., Tornede, A., H{\"u}llermeier, E.: Automating multi-label
  classification extending ml-plan  (2019)

\bibitem{zimmer2021auto}
Zimmer, L., Lindauer, M., Hutter, F.: Auto-pytorch: multi-fidelity metalearning
  for efficient and robust autodl. IEEE PAMI  \textbf{43}(9),  3079--3090
  (2021)

\bibitem{zoph2016neural}
Zoph, B., Le, Q.V.: Neural architecture search with reinforcement learning.
  arXiv preprint arXiv:1611.01578  (2016)

\end{thebibliography}
\end{document}